# Improving Statistical Language Model Performance with Automatically Generated Word Hierarchies

J. McMahon* and F.J. Smith

Department of Computer Science, The Queen's University of Belfast

March 9, 1995

## Abstract

An automatic word classification system has been designed which processes word unigram and bigram frequency statistics extracted from a corpus of natural language utterances. The system implements a binary top-down form of word clustering which employs an average class mutual information metric. Resulting classifications are hierarchical, allowing variable class granularity. Words are represented as *structural tags* — unique $n$-bit numbers the most significant bit-patterns of which incorporate class information. Access to a structural tag immediately provides access to all classification levels for the corresponding word. The classification system has successfully revealed some of the structure of English, from the phonemic to the semantic level. The system has been compared — directly and indirectly — with other recent word classification systems. Class based interpolated language models have been constructed to exploit the extra information supplied by the classifications and some experiments have shown that the new models improve model performance.

## 1 Introduction

Many applications which process natural language can be enhanced by incorporating information about the probabilities of word strings; that is, by using statistical language model information [8, 9, 18, 27]. The quality of these components is often measured by *test set perplexity* [2, 1], in spite of some limitations [44] : $PP = \hat{P}(w_1^N)^{-\frac{1}{N}}$, where there are $N$ words in the word stream. With an arbitrarily chosen standard test set, statistical language models can be compared [5]. This allows researchers to make incremental improvements to the models [25]. Cognitive scientists are interested in those features of automatic word classification which have implications for language acquisition [14, 39].

One common model of language calculates the probability of the $i$th word $w_i$ in a test set by considering the $n$ most recent words $\langle w_{i-n}, w_{i-n+1}, \ldots, w_{i-1}\rangle$, or $\langle w_{i-n}^{i-1}\rangle$ in a more compact notation. The model is finitary (according to the Chomsky hierarchy) and linguistically naïve, but it has the advantage of being easy to construct and its structure allows the application of Markov model theory [36].

Much work has been carried out on word-based $n$-gram models, although there are recognised weaknesses in the paradigm. One such problem concerns the way $n$-grams partition the space of possible word contexts. In estimating the probability of the $i$th word in a word-stream, the model considers all previous word contexts to be identical if and only if they share the same final $n$ words. This simultaneously fails to differentiate some linguistically important contexts and unnecessarily fractures others. For example, if we restrict our consideration to the two previous words in a stream — *i.e.* to the trigram conditional probability estimate $\hat{P}(w_i|w_{i-2}^{i-1})$ — then the sentences

(1a) `The boys eat the sandwiches quickly.`

*Supported by the Department of Education for Northern Ireland and British Telecom Research Labs. Corpora supplied by the Oxford Text Archive, except VODIS which was supplied by British Telecom. Author's contact address, Department of Computer Science, Q.U.B., Belfast BT7 1NN, N. Ireland. Email: J.McMahon@qub.ac.uk

and

(2a) `The cheese in the sandwiches is delicious.`

contain points where the context is inaccurately considered identical. These points occur after the common bigram ⟨`the sandwiches`⟩; we can illustrate the danger of conflating the two sentence contexts by considering the non-sentences

\*(1b) `The boys eat the sandwiches is delicious.`

and

\*(2b) `The cheese in the sandwiches quickly.`

In (1a), we should not be surprised to find an adverb such as ⟨`quickly`⟩, but we would be surprised to find the verb ⟨`is`⟩ as shown in (1b); conversely for (2a) and (2b). In both cases, however, we can construct other sentences which make these words more likely; for example :

(1c) `The way the boys eat the sandwiches is delicious.`

and

(2c) `The cheese in the sandwiches quickly melts.`

We can use results of information theory [43] to quantify operationally our degree of surprise. We have seen how word based trigram models fail to differentiate some obviously different contexts; there are some techniques to alleviate this problem [31, 32] but researchers are constantly confronted with the sparseness of the $n$-gram word segments found in even the largest corpora and with powerful criticisms of the finite-state grammars they use, based on theoretical linguistic grounds [7, 37].

A second weakness of word based language models is their unnecessary fragmentation of contexts. The set of words which are interchangeable with a given word in a sentence such that the newly generated sentence is also well formed is called a paradigmatic set. Consider the word ⟨`boys`⟩ in (1a) above. We can structure our entire vocabulary around this word as a series of layers, each layer of which contains words which share many properties with ⟨`boys`⟩. For example, words like ⟨`lads`⟩ might be positioned in an inner paradigmatic layer. Other words might be placed further away — for example ⟨`children`⟩. A linguistically significant layer around the word ⟨`boys`⟩ is one which contains all plural nouns; at this stage however, intersubstitution can produce sentences which may be difficult to interpret.

If sentences (1a) and (2a) are converted to word class streams

(3) `determiner noun verb determiner noun adverb`

and

(4) `determiner noun preposition determiner noun verb adjective`

respectively, then bigram, trigram and possibly even higher $n$-gram statistics may become available with greater reliability for use as context differentiators (although Sampson [41] suggests that no amount of word-class $n$-grams may be sufficient to characterise natural language fully). In particular, we should be able to use trigram contexts ⟨`verb determiner noun`⟩ and ⟨`preposition determiner noun`⟩ to differentiate reliably between the contexts in (1a) and (2a) at the point where the word ⟨`sandwiches`⟩ is being processed. Of course, this still fails to differentiate the contexts in sentences (1c) and (2c); whilst $n$-gram models of language may never fully model long distance linguistic phenomena, we argue that it is still useful to extend their scope.

In order to make these improvements, we need access to word class information, which is usually obtained in three main ways : firstly, we can use corpora which have been manually tagged by linguistically informed experts [13]; secondly, we can construct automatic part-of-speech taggers and process untagged corpora [26, 15]. This method boasts a high degree of accuracy, although often the construction of the automatic tagger involves a bootstrapping process based on a core corpus which has been manually tagged [10]; the third option is to derive a fully automatic word classification system from untagged corpora. Some advantages of this approach include its applicability to any natural language for which some corpus exists, independent of the degree of development of its grammar, and its parsimonious committment to the machinery of modern linguistics. One disadvantage is that the classes which are derived usually allow no linguistically sensible summarising label to be attached. Research has been carried out recently in this area, often independent of any application to statistical language modelling [20, 16, 38, 3, 24, 35, 40, 30]. The next section contains a presentation of a new automatic word classification algorithm.

## 2  Word Classification and Structural Tags

Most statistical language models which make use of class information do so with a single layer of word classes [4] — often at the level of common linguistic classes (nouns, verbs, *etc.*) [13]. Also, underlying these models is an assumed relationship between the set of word-objects in the vocabulary and some set of class-objects, often defined by professional linguists; deriving a propitious classification system corresponds to searching the space of possible word-to-class mappings. In contrast, we present the *structural tag* representation, where the symbol which represents the word simultaneously represents the classification of that word ([29] makes connections between this and other representations — for example Schütze's category space [42] and, ultimately Saussure's [12] structuralist conception of the sign — and [28] describes its implementation more fully). In our model, each word is represented by an $s$-bit number the most significant bits of which correspond to various levels of classification; given some word which is represented as structural tag $w$, we can gain immediate access to all $s$ levels of classification of that word. Generally, the broader the classification granularity we chose, the more confident we can be about the distribution of classes at that level, but the less information this distribution offers us about next-word prediction.

In many word classification systems, the hierarchy is not explicitly represented, and further processing, often by standard statistical clustering techniques, is required (*e.g.* Elman, Schütze, Brill *et al.*, Finch *et al.*, Hughes *et al.* and Pereira *et al.*). With the structural tag representation, each tag contains classification information which is explicitly represented; the position of that word in class-space can be obtained without reference to the positions of other words. Also, many levels of classification granularity can be made available simultaneously, and the weight which each of these levels can be given in, for example a statistical language model can alter dynamically. Using the structural tag representation, the computational overheads for using class information can be kept to a minimum. Also, it is possible to organise an $n$-gram frequency database so that close structural tags are stored near to each other; this could be exploited to reduce the search space explored in speech recognition systems. For example, if the system is searching for the frequency of a particular noun in an attempt to find the most likely next word, then alternative words should already be nearby in the $n$-gram database. Finally, we note that in the current implementation of the structural tag representation we allow only one tag per orthographic word-form; although many of the current word classification systems do the same, we would prefer a structural tag implementation which models the multi-modal nature of some words more successfully.

Consider sentences (1a) and (2a) again; we would like to construct a clustering algorithm which assigns some unique $s$-bit number to each word in our vocabulary so that the words are distributed according to some approximation of the paradigmatic layering described above — that is, ⟨`boys`⟩ should be close to ⟨`people`⟩ and ⟨`is`⟩ should be close to ⟨`eat`⟩. We would also like semantically related words to cluster, so that, although ⟨`boys`⟩ may be near ⟨`sandwiches`⟩ because both are nouns, ⟨`girls`⟩ should be even closer to ⟨`boys`⟩ because both are human types. In theory, structural tag representations can be dynamically updated — for example, ⟨`bank`⟩ might be close to ⟨`river`⟩ in some contexts and closer to ⟨`money`⟩ in others.

Although we could construct a useful set of structural tags manually [28], we prefer to design an algorithm which builds such a classification. The algorithm is locally optimal and implements a binary form of simulated annealing, which, informally, follows Brown's law of cumulative complexity [6, 45]; it states that, during human development, simple cognitive structures are more likely to be acquired before more complex ones. In our version applied to structural tags, this is equivalent to an algorithm which first treats all words in a corpus as if they belonged to only two classes and then attempts to find the optimal two-class word division; it then assumes that words can belong to any of 4 classes and continues as before; the algorithm finishes when it reaches a classification with $2^s$ classes.

For a given vocabulary, $V$, the mapping $t$ initially translates words into their corresponding unique structural tags. This mapping is constructed by making random word-to-tag assignments. Brown *et al.* have shown that any classification system whose average class mutual information is maximised will lead to class-based language models of lower perplexities.

The mutual information [11] between two events $x$ and $y$ is :

$$I(x, y) = \log \frac{P(x, y)}{P(x)P(x)}$$

It follows that if the two events, $x$ and $y$ stand for the occurrence of certain word classes in a sample, then we can estimate the mutual information between the two classes. In these experiments, we use maximum likelihood probability estimates based on a training corpus. In order to estimate the average class mutual information, for a classification depth of $s$ bits we compute :

$$M_s(t) = \sum_{c_i, c_j} P(c_i, c_j) \times \log \frac{P(c_i, c_j)}{P(c_i)P(c_j)} \qquad (1)$$

where $c_i$ and $c_j$ are word classes and $M_s(t)$ is the average class mutual information for structural tag classification $t$ at bit depth $s$. It follows that the optimal classification, $t^o$ can be found by computing

$$M_s(t^o) = \max_t M_s(t) \qquad (2)$$

No method exists at present which can find the optimal classification, but sub-optimal strategies exist which lead to useful classifications. The sub-optimal strategy used in the current automatic word classification system involves selecting the *locally* optimal structure between $t$ and $t'$, which differ only in their classification of a single word. An initial structure is built by using the computer's pseudo-random number generator to produce a random word hierarchy. Its $M(t)$ value is calculated. Next, another structure, $t'$ is created as a copy of the main one, with a single word moved to a different place in the classification space. Its $M(t')$ value is calculated. This second calculation is repeated for each word in the vocabulary and we keep a record of the transformation which leads to the highest $M(t')$. After an iteration through every word, we select that $t'$ which has the highest $M(t')$ value and re-iterate. With this method, words which at one time are moved to a new region in the classification hierarchy can move back at a later time, if licensed by the mutual information metric. In practice, this does happen. Therefore, each transformation performed by the algorithm is not irreversible within a level, which should allow the algorithm to explore a larger space of possible word classifications.

The algorithm is embedded in a system which calculates the best classifications for all levels : first, the highest classification level is processed. Since the structural tag representation is binary, this first level seeks to find the best distribution of words into two classes.

When the locally optimal two-class hierarchy has been discovered by maximising $M_1(t)$, whatever later re-classifications occur at finer levels of granularity, words will always remain in the level 1 class to which they now belong. For example, if many nouns now belong to class 0 and many verbs to class 1, later sub-classifications will not influence the $M_1(t)$ value. This reasoning also applies to all classes $s = 2, 3 \ldots 16$.

This algorithm, which is $O(sV^3)$ for vocabulary size $V$, works well with the most frequent words from a corpus; however we have developed a second algorithm, to be used with the first, to allow vocabulary coverage in the range of tens of thousands of word types. This second algorithm exploits Zipf's law [46] — the most frequent words account for the majority of word tokens — by adding in low frequency words only after the first algorithm has finished processing high frequency ones. We make the assumption that any influence that these infrequent words have on the first batch of frequent words can be discounted. The algorithm is an order of magnitude less computationally intensive and so can process many more words in a given time. By this method, we can also avoid modelling only a simplified subset of the phenomena in which we are interested and hence avoid the danger of designing systems which do not scale-up adequately [14].

# 3   Word Classification Results

We applied our algorithm to several corpora in order to explore its scope of application. It successfully discovers major noun-verb distinctions in a well-known type 4 grammar introduced by Elman [14], makes near perfect vowel-consonant distinctions when applied to a phonemic corpus and makes important syntactic and semantic distinctions in a Latin corpus; [28] contains full descriptions of these and several other results, including a detailed comparison with a merge-based clustering technique which Brown *et al.* incorporate as part of their word clustering system. We have found that the merge-based techniques implemented by Brown *et al.*, Hughes, Finch *et al.*, Schütze and Brill *et al.* result in taxonomies which have complementary strengths and weaknesses compared to our top-down classification, which leads us to suggest that hybrid top-down and bottom-up systems might inherit the strengths of both approaches. In the present paper however, we report on the performance of our top-down algorithm when applied to the most frequent words from an untagged version of the LOB corpus [23] and also when applied to a hybrid word and class version of the LOB. We use structural tags which are 16 bits long and we considered the 569 most frequent words. With these experiments we aim to illustrate the quality of the clustering technique; we include a summarising performance measure, developed by Hughes *et al.* which, whilst not ideal [28], represents an objective measure of word cluster quality.

In figure 1, we observe the final state of the classification, to a depth of five bits. Many syntactic and some semantic divisions are apparent — prepositions, pronouns, verbs, nouns and determiners cluster — but many more distinctions are revealed when we examine lower levels of the classification. For example, figure 2 shows the sub-cluster of determiners whose initial structural tag is identified by the four-bit schema 0000. In figure 3 we examine the finer detail of a cluster of nouns. Here, some semantic differences become clear. Many of the $2^5$ groups listed in figure 1 show this type of interesting fine detail. In another experiment, a hybrid version of the LOB corpus was created by replacing each word and part-of-speech pair by the word only if the part-of-speech was a singular noun, the base form of a verb, or the third person singular present tense of a verb and by the part-of-speech itself otherwise. When we examine the most frequent 'words' of this hybrid corpus, we find that there are many more content words present, but that the remaining content words still have an indirect effect on word classification, since they are represented by the part-of-speech of which they are an example. Figures 4 and 5 show many of the largest groupings of words found after processing, at a classification level of 9 bits. By inspection, we observe a variety of semantic associations, although there are anomalies. In each group we include here, the entire membership is listed. The remaining groups not presented here also display strong semantic clustering. Finally, figure 6 shows the complete phoneme classification of a phonemic version of the VODIS corpus. The most obvious feature of this figure is the successful distinction between vowels and consonants. Beyond the vowel-consonant distinction, other similarities emerge : vowel sounds with similar vocal tract positions are clustered closely — the ⟨a⟩ sounds, for example, and the phoneme pair ⟨o⟩ and ⟨oo⟩; some consonants which are similarly articulated also map onto local regions of the classification space — ⟨r⟩ and ⟨rx⟩, ⟨ch⟩ and ⟨z⟩ and ⟨n⟩ and ⟨ng⟩, for example.

Assessing the quantitative significance of these results is difficult; Hughes [19], for example, suggests benchmark evaluation — a standard tagged corpus is used as a reference against which automatic comparisons can be made; whilst this may not be appropriate for the designers of every automatic classification system (for example, researchers whose main interest in automatic classification is as a means towards the end of improving statistical language models), it has many advantages over qualitative inspection as an evaluation method, which to date has been the dominant method. The evaluator developed by Hughes allows accuracy scores to be made at each level in a hierarchy. Figure 7 shows performance curves for three systems — that of Finch [17], which uses Spearman's rank correlation coefficient, that of Hughes, which used Ward's method, and the top-down mutual information algorithm. The top-down algorithm fares well, even though it is currently based on contiguous bigram information only, whereas both of the other classification systems are based on contiguous and non-contiguous bigrams — that is, bigrams of the form $\langle w_{i-1}, w_i \rangle$ and $\langle w_{i-2}, w_i \rangle$. On Hughes *et al.*'s evaluation measure, our system performs better that Finch *et al.*'s but worse than Hughes *et al.*'s.

Not only can we estimate and compare the inherent linguistic quality of the classification, but we can also assess its utility by measuring how much it improves the most successful statistical language models to date — those based on Markov model theory. This indirect evaluation allows us to examine the significance of sub-syntactic clusters; for example, even though a high level cluster may contain many nouns, the direct evaluation of Hughes *et al.* is currently unable to reward successful semantic sub-clusterings.

## 4  Structural Tags and Statistical Language Models

There are several ways of incorporating word classification information into statistical language models using the structural tag representation [28]. Here, we shall describe a method, derived from Markov model theory [21], which is based on interpolating several language components. The interpolation parameters are estimated by using a held-out corpus.

For the following experiments, a formatted version of the one million word Brown corpus was used as a source of language data; 60% of the corpus was used to generate maximum likelihood probability estimates, 30% to estimate frequency-dependent interpolation parameters, and the remaining 10% as a test set.

For completeness, we also calculated some test-set perplexities of simple language models. The simplest unigram language model is an equiprobable one; its perplexity is 33,360. A maximum likelihood unigram model, whilst not using held-out data, results in a test set perplexity of 1,226.7. The simplest language model system to use held-out data is a frequency-independent, word-based trigram language model, summarised by the equation :

$$P(w_k) = \lambda_u \times P(w_k) + \lambda_b \times P(w_k|w_j) + \lambda_t \times P(w_k|w_i, w_j)$$

The three parameters $\lambda_u$, $\lambda_b$ and $\lambda_t$, corresponding to unigram, bigram and trigram weights, initailly are given equal weights of $\frac{1}{3}$; a simplified version of the Forward-Backward algorithm [34] is iterated until the halting condition $|\lambda^{t+1} - \lambda^t| < 0.001$. This condition is chosen arbitrarily, but stays constant throughout the remaining experiments. The resulting language model system reduces test set perplexity to 701.7.

Improved performance can be obtained by making interpolation parameters depend upon some distinguishing feature of the prediction context. One easily calculated feature is the frequency of the previously processed word. In our experiment, this resulted in 428 sets of $\lambda$ values. The optimised parameters are fitted into an interpolated language model the core of which is described by the equation :

$$P(w_k) = \lambda_u(f) \times P(w_k) + \lambda_b(f) \times P(w_k|w_j) + \lambda_t(f) \times P(w_k|w_i, w_j)$$

where $f = f(w_j)$, the frequency of word $\langle w_j \rangle$ if there a valid $w_j$ exists and 0 otherwise — namely at the beginning of the test set, and when the previous word is not in the training vocabulary. The resulting perplexity value for this system is 621.6. This represents a pragmatically sensible baseline value against which any variant language model should be compared. Another similar word-based language model has been developed by O'Boyle *et al.* [33], the weighted average language model. This model is described as follows :

$$P(w_k|w_1^{k-1}) = \frac{\sum_{i=1}^{m} \lambda_i \times P_{ML}(w_k|w_{k-i}^{k-1}) + \lambda_0 \times P_{ML}(w_k)}{\sum_{i=0}^{m} \lambda_i}$$

where there are statistically significant segments up to $m+1$ words long and $P_{ML}(w_k)$ is the maximum likelihood probability estimate of a word. The numerator acts as a normaliser. It has been found that

$$\lambda_i = 2^{(|w_{k-i}^{k-1}|)} \times \log f(w_{k-i}^{k-1})$$

where $|w_{k-i}^{k-1}|$ is the size of the segment, results in a near-optimum language model of this form. When applied to the Brown corpus, excluding the 30% which has been allocated for interpolation, the model still performs well, achieving a perplexity score of 630.9, 1.5% above the interpolated optimum value.

Using structural tags which are 16 bits long allows us, in principle, to build an interpolated model which contains 16 levels of word class information. The model can be described as follows :

$$P(w_k) = \sum_{s=1}^{16} \sum_{i=1}^{3} \lambda_i^s(f) \times P_i^s(w_k)$$

where $i = 1, 2, 3$ corresponds to unigram, bigram and trigram components respectively, and $P_i^s(w_k)$ is the probability estimate of the $i$th language model of depth $s$ for word $w_k$. For example, if we are processing the word ⟨sandwiches⟩ in the sequence ⟨eat the sandwiches⟩, then $P_2^7$(sandwiches) is the estimate of the bigram conditional probability $P(c^7(\text{the sandwiches})|c^7(\text{the})) \times P(\text{sandwiches}|c^7(\text{sandwiches}))$, where $c^7$(the sandwiches) is the structural tag equivalent of ⟨determiner noun-plural⟩ — that is, the representation of that class bigram, at depth 7, the first element of which contains as a class member the word ⟨the⟩ and the second element of which contains as a class member the word ⟨sandwiches⟩. If our word classification system performs well, we should expect other foodstuff words to be included in the $c^7$ classification. Figure 3 lends support to our belief that our algorithm discovers interesting word classes — it illustrates a $c^9$ group which contains the words {boy girl child man woman person}. Figures 4 and 5 contain many semantic groupings observed at the $c^9$ level.

Due to the modest size of the Brown corpus, we did not implement a full-level interpolated class-based language model — there would have been too many training parameters and not enough training data; instead, we implemented several two-level and three-level variants of the system, the most successful of which is a three-level system which takes information from level 16 (corresponding to a standard word-based trigram system), level 8 and level 5. This language model registered a test set perplexity of 586.5, compared to the trigram word-based language model score of 621.6. Figure 8 summarises all of the test set perplexity results.

## 5 Discussion

As an illustration of the kind of advantage which structural tag language models can offer, we introduce nine oronyms based upon the uttered sentence

> the boys eat the sandwiches

If we assume that we already possess a perfect speech recognition acoustic model [22], it may be able to recover the optimal phoneme string :

> /DH a b OI z EE t DH A s AA n d w i j i z/

However, the original sentence is not the only speech utterance which could give rise to the observed phoneme string. For example, the meaningless and ungrammatical sentence

> the buoy seat this and which is

can also give rise to the observed phonemic stream. Humans usually re-construct the most likely sentence successfully, but artificial speech recognisers with no language model component cannot. A useful statistical language model will assign a low probability to the second sentence and a high probability to the first. A more traditional, non-probabilistic language model, in the form of a grammar, could also differentiate between the two sentences, weeding out the second, ungrammatical sentence. However, such models, whilst theoretically well grounded, so far tend to have poor coverage. Another problem with non-probabilistic models can be seen if we consider a third hypothesised sentence :

> the buoys eat the sand which is

This simultaneously surreal and metaphysical sentence may be accepted by grammar systems which detect well-formedness, but it is subsequently considered just as plausible as the original sentence. A probabilistic language model should assign a relatively low probability to the third sentence. We constructed nine hypothesised sentences, each of which could have produced the phoneme string; we presented these sentences as input to a high-quality word-based language model (the weighted average language model) and to the 16-8-5 three-level structural tag language model. Figure 9 shows the normalised probability results of these experiments. The new language model successfully identifies the most likely utterance. In all but two cases, sentences which are grammatically well-formed are assigned a higher raw probability by the new model, and vice-versa for ungrammatical sentences. Using the top two sentences ⟨the boy seat the sandwiches⟩ and ⟨the boys eat the sandwiches⟩, we can examine the practical benefits of class information for statistical language modelling. The only difference between the two is in the bigrams ⟨boy seat⟩ and ⟨boys eat⟩, neither of which occured in the training corpus. For the model which uses word frequencies exclusively, it differentiates between the two hypothesised sentences by examining the unigrams ⟨boy⟩, ⟨seat⟩, ⟨boys⟩ and ⟨eat⟩. In our training corpus, ⟨boy⟩ and ⟨seat⟩ are individually more likely than ⟨boys⟩ and ⟨eat⟩. However, with the 16-8-5 structural tag model, extra word class information allows the system to recognise the familiar `noun-verb` pattern as being more likely than the `noun-noun` pattern.

The automatic word classification system based on a binary top-down mutual information algorithm leads to qualitatively impressive syntactic and semantic clustering results (see figures 1 to 5); quantitatively, it fares well with other successful systems, demonstrating complementary strengths and weaknesses compared to the more usual merge-based classification systems, even though it currently uses contiguous bigrams (see figure 7 and also [28] for a direct comparison with another merge-based classification system); thirdly, results from a simple implementation of the system (3 classification levels out of 16) show a significant improvement in statistical language model performance. We have incorporated structural tag information into an interpolated trigram language model system because it provided a well-attested and successful base system against which we could measure improvement; however, we believe that the weighted average system described earlier, with its scope for including $n$-gram information beyond the trigram and its avoidance of data-intensive and computationally intensive parameter optimisation, could offer a more convenient platform within which to place structural tag information. We believe that although variable granularity class-based language models may never fully capture linguistic dependencies, they can offer modest advances in coverage compared to exclusively word-based systems.

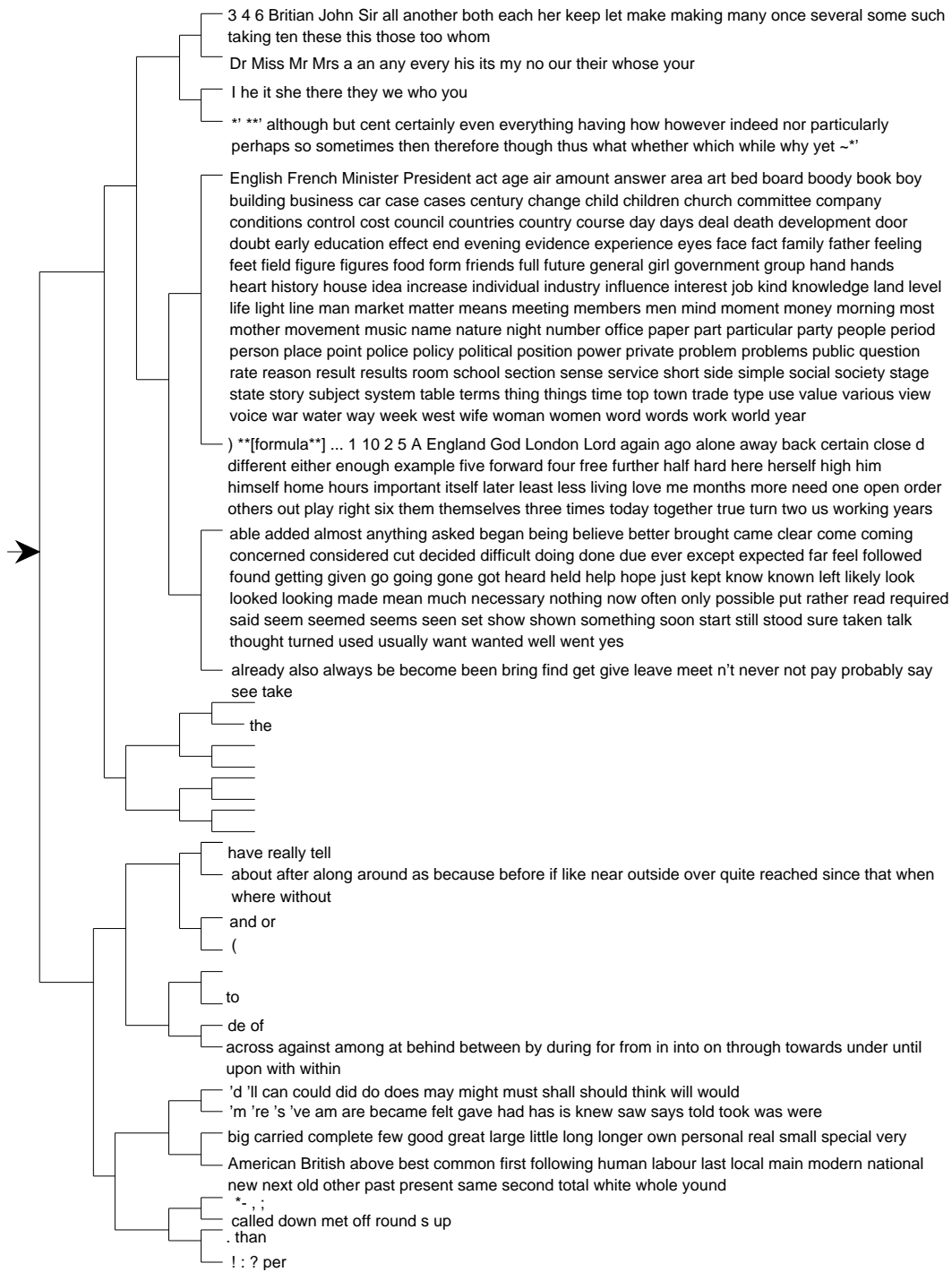

Figure 1: Final distribution of the most frequent words from the LOB corpus. Only the first five levels of classification are given here, but important syntactic relations are already clear.

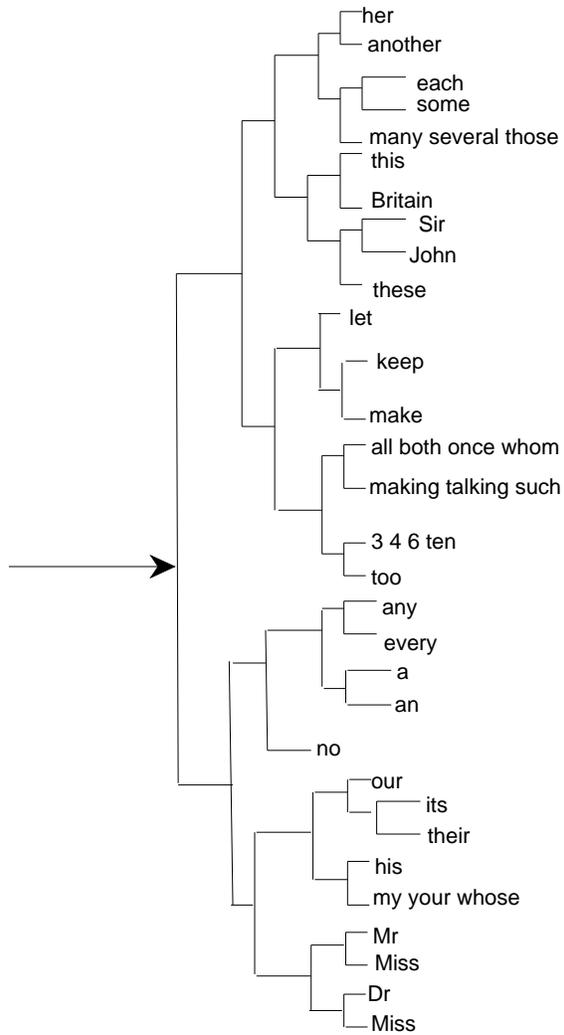

Figure 2: Detail of relationship between words whose final tag value starts with the four bits 0000. Many of these words exhibit determiner-like behaviour.

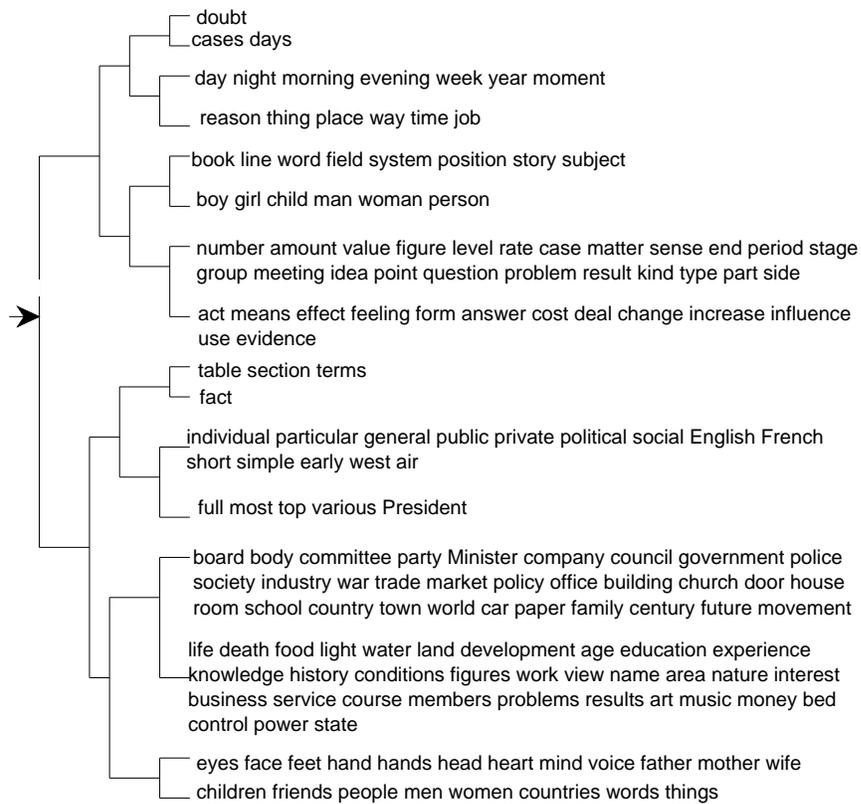

Figure 3: Detail, from level 5 to level 9, of many noun-like words. Clear semantic differences are registered.

> arm breath breathing  cheek chin coat eye fist forehead hair handkerchief hat head knee lad memory mouth neck nose purse shoulder skirt throat whistle wrist ability curiosity destination discretion employer hairdresser heritage inheritance intention tone irritation lifetime loneliness midst mistress profession reluctance typewriter career castle

> aunt brother father father-in-law husband mother sister uncle wife gaze jaw lip voice diary wallet companion lordship partner wholesaler

> approval charm conscience contempt disposal embarrassment engagement fate fault good ideal imagination mind name opinion resignation sake soul speech temper thesis vision will arrival audience correspondent tutor designer friend neighbour lord lover environment childhood cousin daughter son niece elbow hand tongue lap apartment cave surgery uniform

> decade year month fortnight week hour inch lot spot step string host row pool pot ending matter instant minute moment while pause second bit minimum way clue chance fool illusion joke gasp mistake nuisance offence pity reason proposition enquiry job ship compartment room chair leg ridge blanket altar envelope powder adult debt determination disposition

> actor actress artist boy bride captain catholic chap child citizen composer couple critic doctor engineer fellow gentleman girl god hostess individual journalist king lady lawyer legend man novelist observer painter patient people person priest prisoner producer psychologist queen ruler scholar scientist singer soldier sovereign stranger teacher widow woman writer bird cow creature dog mantis rat tree assembly republic accident incident situation experiment target corruption scandal struggle armistice invitation obligation opportunity temptation tendency willingness desire passion mood fortune multitude maximum alternative amendment proposal decision document lease catechism inscription phrase word poem attempt effort gesture sigh whisper message bell knife pen meal poultice clock taxi cloud lawn staircase device explosive explosion shot ray shower dress cathedral hut

> living-room lounge attic roof bathroom bedroom dining-room drawing reception drawing-room kitchen sitting-room carpet floor window cooker oven cabinet desk lamp mirror cupboard shelf telephone sofa hearth fire fireplace flame turf corner corridor hall door doorway entrance tunnel lock key gate background rear boundary hedge wall yard ground surface square garage garden terrace farm farmhouse outside field courtyard barn stable cabin cage pit cell cottage flat studio hotel house laboratory palace tower bay beach lake pond river pier mud net city town village neighbourhood country nation landscape park forest hill slope mine lane road street island moon moonlight sun sky storm wind weather horse tractor continent world landlord owner maid manufacturer peasant cafe clinic concert funeral inquest hearing honeymoon wedding airport platform helicopter boat bus car coach plane flight journey mission model bishop vicar parish church pulpit throne title temple devil gospel truth oath trio tribe budget loan pension penalty prize exchequer taxpayer campaign presentation election term conservative ambulance wound brain baby body finger chest waist stomach womb heart flesh skin heel collar pocket bomb receiver tape gun pistol rifle bullet sergeant drillbriefcase dictionary magazine camera cigarette jacket bow shaft stem sword wing ontrary craft crying dark week-end weekend evening future past defendant registrar spectator student subject editor grandfather downwash flood trawl incidence left reverse outset peak waltz wolf nucleus

> feels regards sees thinks knows likes loves wants wishes seeks calls asks changes describes says uses plays owns loses

> admit assume believe think understand realise realize reckon conclude remember decide dare deserve expect suppose seem tend appear prefer want own bother afford refuse fail feel forget hate hesitate intend know learn like  begin propose try continue cease

> acts arises begins consists depends lies occurs refers rests varies

> assumption certainty doubt hope wonder

> endeavour excuse hurry need request risk urge right sign

Figure 4: Semantic clustering results. Memberships of 11 of the 38 classes whose size is greater than ten, at a classification level of 9 bits. From the LOB corpus. Body parts, relatives, mental states, human roles, house parts, two classes of mental verb, relation verbs, hope-nouns, effort verbs.

admiration affection agony anger ambition amusement anticipation anxiety apathy authority awareness beauty belief bitterness chaos character coincidence communication composition continuity concentration concern confidence conflict conformity controversy confusion consciousness courage courtesy criticism cruelty depression despair destiny devotion difficulty dignity disappointment disorder disturbance drama efficiency enjoyment entertainment enthusiasm equality evil excitement experience failure fairness fantasy freedom friendship frustration fun genius goodwill gossip grace gratitude gravity grief growth guilt happiness harmony haste hatred hospitality humanity humour ideology ignorance impatience impulse inclination independence indignation injustice inspiration interest jealousy joy judgement judgment justice kindness laughter learning liberty licence luck maturity morality nonsense optimism pain patience perception personality perspective philosophy pleasure popularity pregnancy prejudice prestige pride principle quality questioning rage realism relief resentment respect responsibility rivalry romance salvation satisfaction scholarship schooling shame shock silence sin size skill stability style success symmetry sympathy talent temperament tension terror tiredness tragedy triumph trouble uncertainty urgency victory vigour violence virtue warmth weakness wealth weight spirit communism democracy socialism federation parliament technology tradition revolution evolution progress nature society literature poetry knowledge thinking singing writing thought language logic mathematics disarmament politics economics economy investment inflation employment productivity policy pornography poverty publicity reality religion hell paradise theology action effect activity routine output production occupation motion movement operation occurrence behaviour capacity essence shape being baptism life death suicide punishment execution disease illness injury medicine destruction disaster imprisonment residence settlement adultery sex marriage crime warning error myth after-care agriculture admission access exposure convenience addition accumulation expansion consumption possession excess completion duration consequence contamination contact greeting assistance evidence information example selection explanation observation interpretation instruction behalf detail attention consideration conversation consultation permission investigation inspection supervision view visibility sight discussion advantage celebration ceremony tribute accordance preparation recognition protection redemption remuneration provision pursuit adjustment approximation limitation precision compromise negotiation analogy correspondence comparison parallel practice contrast response reference relation rhythm conjunction connection connexion integration implication suspicion rumour outlook direction expulsion acid antibody radioactivity oxidation chemistry physics dose part thread bottle sweat fluid liquid gin sherry whisky wine bream cap cream toast gear material money payment hay vegetation wheat wood wool rose acre pitch place hole layer length age date midnight instance annum white black mist chairman president decoration painting firing noise sport appendix diagram fig p Op page paragraph momentum heaven suet aluminium cotton gold hydrogen oxygen polythene sodium grill salvage football tennis prison wartime grading boiling baking cooking engineering manufacturing shopping housing boarding trading export finance finishing sintering electricity ignition motor rocket railway science-fiction bismuth brass metal nerve oak resin grammar noun infant kinship dairy census county amateur golf bending torsion-flexure drift equilibrium equity rotor coolant ion exhaust humus leather plastic textile consumer retail labour wage borstal constituency qualifying slenderness bend boil brush clasp cover cut feed fly grasp help hurt kick kiss knock melt move offset open pass pop push register repair repeat ride ruin run sail scream sleep stand start stay stick swim talk tie turn wake walk wash watch wear worry blame regret call concentrate look count fancy offer guess mention

Figure 5: More semantic clustering results. Memberships of 6 of the 38 classes whose size is greater than ten, at a classification level of 9 bits. From the LOB corpus. Mental states, writing reference, materials, materials and processing, more materials and processing, common manipulation verbs.

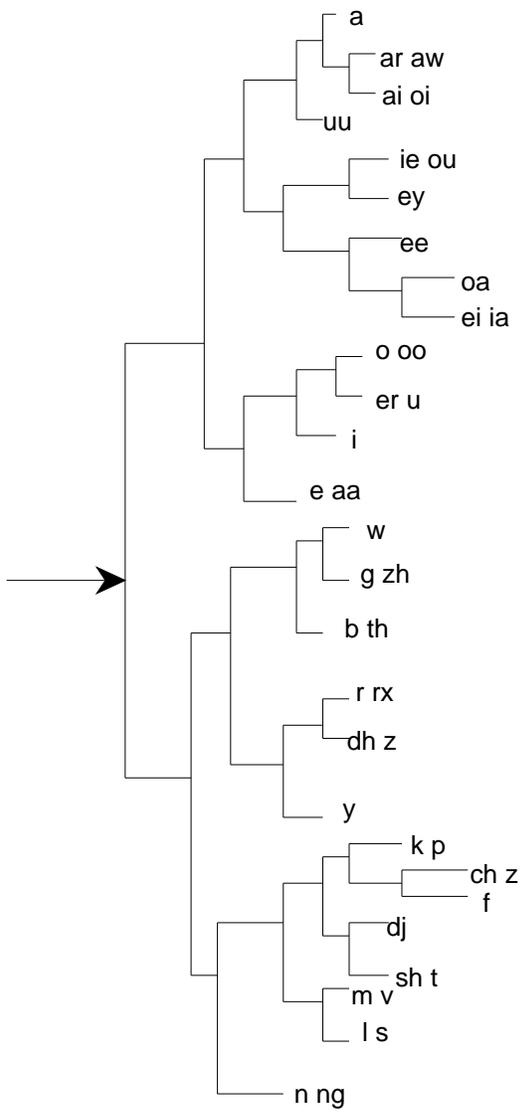

Figure 6: Automatic Phoneme Clustering; differentiation between vowels and consonants. From a phonemic version of the VODIS corpus.

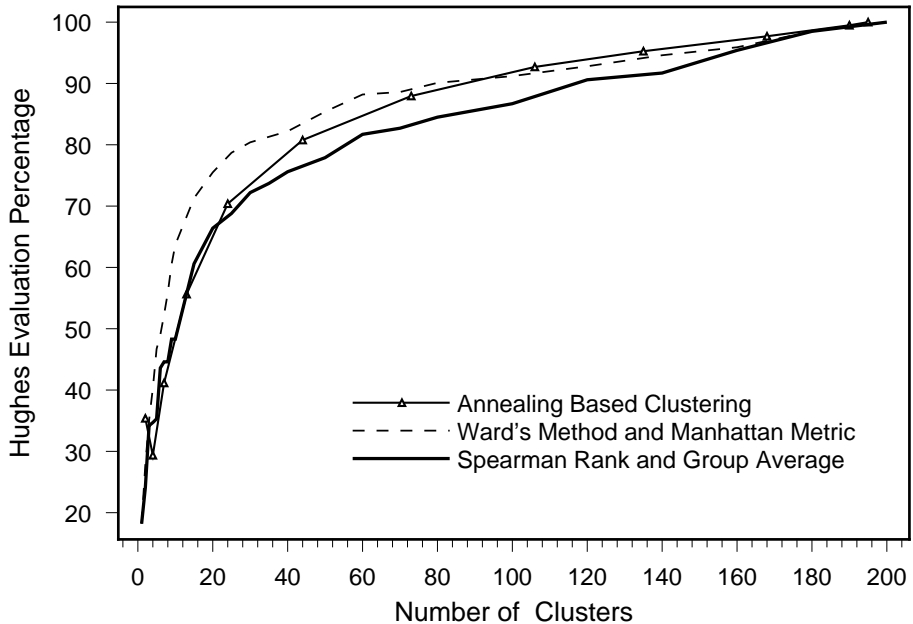

Figure 7: Graph showing the performance of our annealing classification system compared to two recent systems — those of Hughes and Atwell and Finch and Chater. Performance is measured by the Hughes-Atwell cluster evaluation system using the tagged LOB corpus.

| Language Model | Test Set Perplexity |
|---|---:|
| Equiprobable | 33,360 |
| Unigram | 1,226.7 |
| Interpolated Trigram ($f$-independent) | 701.7 |
| Weighted Average | 630.9 |
| Interpolated Trigram ($f$-dependent) | 621.6 |
| 16-8-5 Structural Tag | 586.5 |

Figure 8: A list of language model systems, which shows a steady reduction in test set perplexity. The weighted average model result is based on two thirds of the training material available to the other systems. A three-level class-based model has results in an improvement over the frequency-dependent interpolated trigram model.

| sentence | W.A. | 16-8-5 |
|---|---|---|
| the boy seat the sandwiches | 0.55970 | 0.44966 |
| the boys eat the sandwiches | 0.29255 | 0.50538 |
| the boy seat this and which is | 0.07128 | 0.00783 |
| the boys eat this and which is | 0.03794 | 0.00855 |
| the buoys eat the sandwiches | 0.03189 | 0.02688 |
| the buoys eat this and which is | 0.00414 | 0.00045 |
| the boys eat the sand which is | 0.00225 | 0.00119 |
| the buoys eat the sand which is | 0.00024 | 0.00006 |
| the buoy seat this and which is | 0 | 0 |

Figure 9: Nine versions of a phonemically identical oronym, ordered by weighted average (W.A.) probability, normalised over the test sequences. The W.A. language model ranks the preferred sentence second. The 16-8-5 structural tag model successfully predicts the original utterance as the most likely.